\documentclass[conference]{IEEEtran}
\usepackage{amsmath}  
\usepackage{amssymb}  
\usepackage{graphicx} 
\usepackage{subcaption} 
\usepackage{hyperref} 
\usepackage{booktabs} 
\usepackage{pifont}
\usepackage{marvosym}
\usepackage{amssymb}  

\IEEEoverridecommandlockouts
\usepackage{cite}
\usepackage{amsmath,amssymb,amsfonts}
\usepackage{algorithmic}
\usepackage{graphicx}
\usepackage{textcomp}
\usepackage{xcolor}
\def\BibTeX{{\rm B\kern-.05em{\sc i\kern-.025em b}\kern-.08em
    T\kern-.1667em\lower.7ex\hbox{E}\kern-.125emX}}

\begin{document}
\title{NeuroPathNet: Dynamic Path Trajectory Learning for Brain Functional Connectivity Analysis 
}

\author{
\IEEEauthorblockN{1\textsuperscript{st} Tianqi Guo}
\IEEEauthorblockA{\textit{Sussex Artificial Intelligence Institute} \\
\textit{Zhejiang GongShang University}\\
Hangzhou, China \\
tg341@sussex.ac.uk}
\and
\IEEEauthorblockN{2\textsuperscript{nd} Liping Chen}
\IEEEauthorblockA{\textit{School of Computing Technologies} \\
\textit{RMIT University}\\
Melbourne, Australia \\
lp.chen@ieee.org}
\and
\IEEEauthorblockN{3\textsuperscript{rd} Ciyuan Peng}
\IEEEauthorblockA{\textit{Institute of Innovation, Science and Sustainability} \\
\textit{Federation University Australia}\\
Ballarat, Australia \\
ciyuan.p@ieee.org}
\and
\IEEEauthorblockN{4\textsuperscript{th} Jingjing Zhou}
\IEEEauthorblockA{\textit{School of Information and Electronic Engineering} \\
\textit{Zhejiang GongShang University}\\
Hangzhou, China \\
zhoujingjing@zjgsu.edu.cn}
\and
\IEEEauthorblockN{5\textsuperscript{th} Jing Ren\textsuperscript{(\Letter)}}
\IEEEauthorblockA{\textit{School of Computing Technologies} \\
\textit{RMIT University}\\
Melbourne, Australia \\
jing.ren@ieee.org}
}

%
%
%


\maketitle

\begin{abstract}
Understanding the evolution of brain functional networks over time is of great significance for the analysis of cognitive mechanisms and the diagnosis of neurological diseases. Existing methods often have difficulty in capturing the temporal evolution characteristics of connections between specific functional communities. To this end, this paper proposes a new path-level trajectory modeling framework (NeuroPathNet) to characterize the dynamic behavior of connection pathways between brain functional partitions. Based on medically supported static partitioning schemes (such as Yeo and Smith ICA), we extract the time series of connection strengths between each pair of functional partitions and model them using a temporal neural network. We validate the model performance on three public functional Magnetic Resonance Imaging (fMRI) datasets, and the results show that it outperforms existing mainstream methods in multiple indicators. This study can promote the development of dynamic graph learning methods for brain network analysis, and provide possible clinical applications for the diagnosis of neurological diseases. The source code is available at: \url{https://github.com/zhaoxiuye123/NeuroPathNet}.
\end{abstract}


%
\IEEEpeerreviewmaketitle

\section{Introduction}
\label{sec:introduction}

As one of the most complex and dynamic systems in nature, the human brain can be partitioned into multiple regions based on functional coordination.
Recent developments in brain network analysis have transitioned from classical statistical models to increasingly sophisticated deep learning-based methods~\cite{Ji2023SurveyECN,ren2024brain}. Functional magnetic resonance imaging (fMRI) is an effective tool for studying the brain. By detecting blood oxygen level dependent (BOLD) signals, fMRI can provide researchers with rich four-dimensional spatiotemporal data~\cite{biswal2010discovery}, helping researchers observe brain activity patterns at different time scales and better understand the interactions among functional brain communities. 


Despite notable advancements in brain network analysis, existing methods still face critical limitations. Traditional brain network modeling methods usually regard human brain as a static system, ignoring the characteristics of the dynamic evolution of neural activity over time~\cite{Cui2023BrainGB,Wang2021GCNfMRI}, and it is difficult to explain the changes in the connection relationship between different functional communities over time. Although some studies have tried to introduce the dynamic functional connectivity mechanism to model temporal characteristics~\cite{Alonso2025FCdynamic}, most of these methods remain at the full graph level, unable to systematically capture the dynamic behavior of specific functional communities, and lack structured and interpretable modeling methods.

To tackle the aforementioned limitations, we introduce NeuroPathNet, a novel framework for brain network representation learning. Inspired by path-level trajectory modeling, this framework captures temporal dynamics from the connectivity pathways among functional brain regions. Specifically, NeuroPathNet treats the functional connections between each pair of brain communities as independent path trajectories, each of which is modeled by a temporal neural network to learn its unique dynamic evolution. These trajectory-level representations are then integrated through a global aggregation module, enabling the joint analysis of multiple functional pathways and the generation of a discriminative whole-brain representation.

Our main contributions can be summarized as follows:

\begin{enumerate}
\item This paper aims to model the temporal evolution of the connectivity between brain functional communities and proposes a pathway-level modeling framework to characterize the dynamic changes of brain community path trajectories.

\item We design a graph neural network framework, NeuroPathNet, with time series modeling and path aggregation capabilities. For the first time, we achieved modeling of dynamic interaction patterns of brain regions at the functional connection path level, and improved the global information integration capability.

\item  We evaluate NeuroPathNet on multiple neuroimaging datasets, showing consistently strong performance across both pathological and normal cases.
\end{enumerate}

\section{Related Work}

\subsection{Graph Transformer Models}

In recent years, graph-based models have achieved good results in various graph learning tasks. These models show excellent adaptability to various downstream tasks of graph learning~\cite{damianou2024graph,ren2023graph}. Transformers have become increasingly important in building graph-based models due to their unique advantages (parameter-rich architecture, powerful learning ability)~\cite{shehzad2024graph}. Recent advances in graph transformer architectures are able to address various aspects of graph learning. For example, Polynormer~\cite{abs_2403_01232} introduced a breakthrough method that processes large-scale datasets through polynomial feature learning. To improve the efficiency of graph learning processing, Graphormer~\cite{ying2021transformers} integrates multiple embedding types to learn complex graph structure information. Based on Graph Transformers, Spectral Attention Networks (SAN)~\cite{kreuzer2021rethinking} introduce a new type of learnable position encoding method.

Contributions in this field also include GTN~\cite{yun2022graph}, which combines path generation with the Transformer architecture of heterogeneous graphs. At the same time, Graph Transformer\cite{dwivedi2020generalization} extends the scope of application to arbitrary graphs by incorporating Laplacian eigenvectors and edge information into its attention mechanism.
Recent developments in Graph Transformers have focused on addressing specific challenges in graph learning. Luo \textit{et al.}\cite{luo2024fairgt} explored the effectiveness and fairness of applying high-order features in Transformers to improve node classification tasks.

\subsection{Brain Network Analysis}
With the advanced neuroimaging technology and neuroscience computational methods, brain network analysis has become a key field to understand brain graph interactions ~\cite{JiZLYZS23}. The effectiveness of graph learning makes it a common method for brain network analysis~\cite{li2021braingnn}.
Various studies have shown significant progress in this field. SMGC~\cite{GaoLLZLLHZWYLC24} developed a new method to identify directions in resting-state networks, which helps us gain insight into the relationships between brain regions. BrainGNN~\cite{li2021braingnn} uses an innovative framework to convert fMRI signals into functional connection networks by treating the brains of different individuals as different graph structures. Although graph transformers and their applications in various fields have made significant progress, they still face significant challenges when applied to brain networks. Developing graph transformers that can effectively model the dynamic communities of the brain and apply them to brain science remains an important but unresolved challenge.

\section{Methodology}

\subsection{Overall Framework}
\begin{figure}[t!]
    \centering
    \includegraphics[width=\columnwidth]{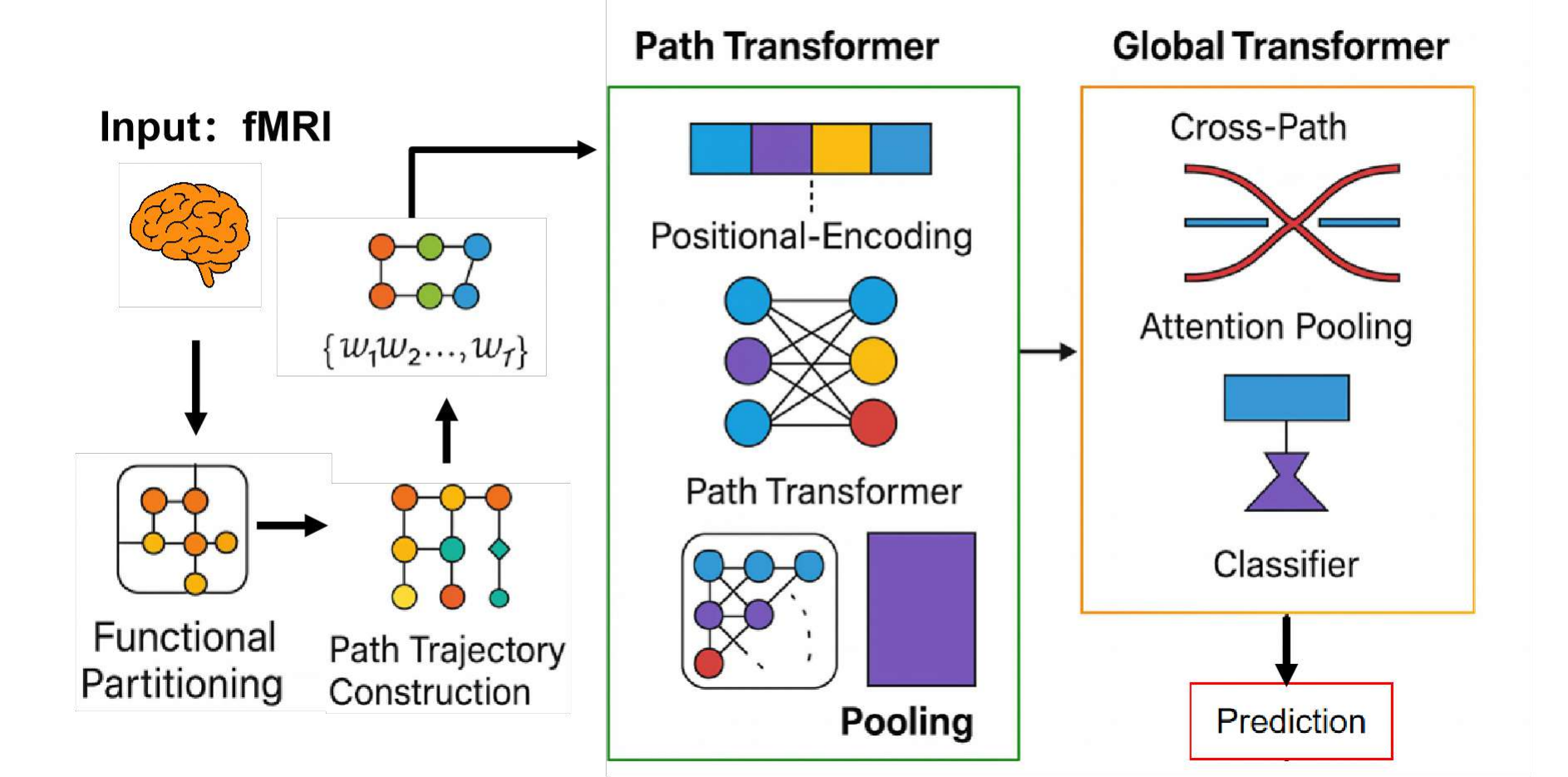}
    \caption{The model takes fMRI data as input, partitions the brain into functional modules, and computes inter-module correlations over time to construct dynamic path sequences. Each path, encoded as a time-series vector with positional information, is fed into a Path Transformer to model temporal interactions. Cross-path attention then integrates all path representations, followed by attention pooling and classifier to produce the final prediction.}
    \label{fig:brain_network}
\end{figure}

This study proposes a pathway-level modeling framework based on static functional partitioning to characterize the dynamic interaction process between brain functional modules. Fig.~\ref{fig:brain_network} shows the overall framework.
The overall process is as follows: First, according to the static functional partitioning scheme recognized by the neuroscience community (such as Yeo-7\cite{yeo2011organization}); then, the region of interest (ROI) is aggregated into several functional modules to construct a dynamic functional connection sequence by applying the sliding window method to the preprocessed fMRI data; then, the average connection strength between any two partitions is calculated in each time window, and the path trajectory is formed in the time dimension; finally, the temporal modeling method is used to extract features for each pathway, and a global representation is generated through the pathway fusion module for downstream classification tasks.

\subsection{Path Trajectory Construction}
After obtaining the static functional partitions, we further construct the dynamic connectivity relationships between these functional partitions. Since the functional connectivity of the brain exhibits significant time-varying characteristics, the interaction relationships between different functional modules dynamically evolve with changes in cognitive tasks, mental states, or disease conditions. A single static connectivity graph cannot effectively capture this dynamic process\cite{Seeburger2024TimeVaryingFC}.

To address this, we employ the Sliding Window method to divide the complete fMRI time series into $T$ overlapping or non-overlapping time windows. In each time window $t \in \{1, 2, \ldots, T\}$, we calculate the average connectivity strength between any two functional partitions $C_i$ and $C_j$ based on the correlations of the ROIs within the partitions, thereby obtaining a time series of community-level dynamic connectivity graphs.

Let the set of functional partitions be:
\begin{equation}
\mathcal{C} = \{ C_1, C_2, \ldots, C_N \}.
\label{eq:partitions}
\end{equation}
Then, the connectivity weight for each community pair $(C_i, C_j)$ in time window $t$ is defined as:
\begin{equation}             
w_{i,j}^{(t)} = \frac{1}{|C_i||C_j|} \sum_{p \in C_i} \sum_{q \in C_j} r_{p,q}^{(t)},
\label{eq:connectivity}
\end{equation}
where $r_{p,q}^{(t)}$ represents the Pearson correlation coefficient between ROI $p$ and ROI $q$ in the $t$-th time window, and $|C_i|$ denotes the number of ROIs in community $C_i$.

By arranging these connectivity strength values along the time axis, we obtain the path-level connection trajectory for community pair $(C_i, C_j)$, defined as:
\begin{equation}
\text{Path}_{i,j} = \{ w_{i,j}^{(1)}, w_{i,j}^{(2)}, \ldots, w_{i,j}^{(T)} \}.
\label{eq:path}
\end{equation}
This trajectory reflects the dynamic evolution of functional interactions between partitions $C_i$ and $C_j$ throughout the scanning process. Each path can be regarded as a time-series representation of a "functional communication pathway" in the brain,  subsequent models will use these paths as basic modeling units.

\subsection{Path-Level Representation Learning}
\begin{equation}
\mathbf{X}_{i,j} = [\mathbf{x}_1, \mathbf{x}_2, \ldots, \mathbf{x}_T]^\top \in \mathbb{R}^{T \times d},
\label{eq:embedding}
\end{equation}
where $\mathbf{x}_t = \text{PE}(w_{i,j}^{(t)})$ represents the embedding feature at the $t$-th timestep, and $\text{PE}(\cdot)$ denotes the positional encoding function used to incorporate temporal order information.

Subsequently, we employ an improved Transformer encoder\cite{vaswani2017attention} to model this path sequence. First, we compute the query, key, and value matrices:
\begin{equation}
\mathbf{Q} = \mathbf{X}_{i,j}\mathbf{W}_Q, \quad
\mathbf{K} = \mathbf{X}_{i,j}\mathbf{W}_K, \quad
\mathbf{V} = \mathbf{X}_{i,j}\mathbf{W}_V,
\label{eq:qkv}
\end{equation}
where $\mathbf{W}_Q, \mathbf{W}_K, \mathbf{W}_V \in \mathbb{R}^{d \times d_h}$ are learnable parameters, and $d_h$ is the attention dimension. Based on the scaled dot-product attention mechanism, we compute the self-attention output:
\begin{equation}
\text{Attn}(\mathbf{Q}, \mathbf{K}, \mathbf{V}) = \text{softmax}\left(\frac{\mathbf{Q}\mathbf{K}^\top}{\sqrt{d_h}}\right)\mathbf{V}.
\label{eq:attention}
\end{equation}
Building upon this, we enhance the representation capability using multi-head attention and introduce residual connections with layer normalization :
\begin{equation}
\mathbf{Z}_{i,j} = \text{LayerNorm}\left(\mathbf{X}_{i,j} + \text{MultiHead}(\mathbf{Q}, \mathbf{K}, \mathbf{V})\right).
\label{eq:residual_multihead}
\end{equation}
To further compress the temporal dimension information, we apply a learnable temporal pooling module on the output sequence, mapping $\mathbf{Z}_{i,j}$ to a fixed-length path representation vector:
\begin{equation}
\mathbf{h}_{i,j} = \text{TemporalPool}(\mathbf{Z}_{i,j}) \in \mathbb{R}^{d}.
\label{eq:temporal_pool}
\end{equation}
Finally, we input the representation $\mathbf{h}_{i,j}$ of each path $(C_i, C_j)$ into subsequent global interaction modules or classifiers to capture the dynamic coordination patterns between different functional regions in the brain network.

\subsection{Whole-Graph Fusion and Classification}

After obtaining the representation vectors $\mathbf{h}_{i,j}$ for all path pairs $(C_i, C_j)$, we organize them into a complete brain graph representation at the path level:
\begin{equation}
\mathcal{H} = \{ \mathbf{h}_{i,j} \mid 1 \le i < j \le N \},
\label{eq:path_set}
\end{equation}
where $N$ is the number of functional partitions, and $\mathcal{H}$ contains $\binom{N}{2}$ path vectors. To model the global coordination features among these paths, we introduce a Cross-Path Attention mechanism as the whole-graph integration module.

First, we stack all path representations into an input matrix:
\begin{equation}
\mathbf{H} \in \mathbb{R}^{L \times d}, \quad \text{where } L = \binom{N}{2}.
\label{eq:H_matrix}
\end{equation}
We then use a lightweight Transformer structure to model $\mathbf{H}$, capturing global interaction features between paths, and output the fused representation matrix $\mathbf{H}'$:
\begin{equation}
\mathbf{H}' = \text{Transformer}_{\text{global}}(\mathbf{H}).
\label{eq:global_transformer}
\end{equation}
Next, we apply an attention pooling mechanism to $\mathbf{H}'$, compressing all path information into a unified brain graph representation vector:
\begin{equation}
\mathbf{z} = \sum_{l=1}^{L} \alpha_l \cdot \mathbf{H}'_l, \quad 
\alpha_l = \frac{\exp(\mathbf{u}^\top \tanh(\mathbf{W}\mathbf{H}'_l))}{\sum_{k=1}^{L} \exp(\mathbf{u}^\top \tanh(\mathbf{W}\mathbf{H}'_k))},
\label{eq:attention_pooling}
\end{equation}
where $\mathbf{W}$ and $\mathbf{u}$ are learnable parameters. Finally, we feed the global representation $\mathbf{z}$ into a classifier for diagnostic prediction.
\begin{equation}
\hat{y} = \text{Softmax}(\mathbf{W}_{\text{cls}} \cdot \mathbf{z} + \mathbf{b}),
\label{eq:classifier}
\end{equation}
and optimize the model using cross-entropy loss:
\begin{equation}
\mathcal{L}_{\text{CE}} = - \sum_{c=1}^{C} y_c \log(\hat{y}_c),
\label{eq:cross_entropy}
\end{equation}
where $C$ is the number of classes, and $y$ is the one-hot encoded ground truth label.

\section{Experiments}
\subsection{Experimental Setup} 
\subsubsection{Datasets}
We evaluate our model on three publicly available neuroimaging datasets.   \textbf{ADNI}\cite{jack2008alzheimer} (Alzheimer's Disease Neuroimaging Initiative) focuses on studying Alzheimer's disease. The participants were divided into three groups: a normal group  (211 people), a mild cognitive impairment group (195 people), and an Alzheimer's disease diagnosis group (54 people). 
The Autism Brain Imaging Data Exchange (\textbf{ABIDE}) \cite{diMartino2014autism} dataset collects a large amount of neuroimaging data from patients with autism spectrum disorder (ASD) and typically developing (TD) controls. For this study, we analyzed data from 945 participants: 479 ASD patients and 466 TD controls.  \textbf{HCP}\cite{vanessen2012human} dataset, part of the Human Connectome Project's Connectomes Related to Human Diseases initiative, focuses on adolescent anxiety and depression. It includes data from 215 adolescents aged 14–17, with 152 diagnosed with anxiety and/or depressive disorders.

\subsubsection{Baselines}

To evaluate the effectiveness of our proposed model, we selected a comprehensive set of baseline methods, including both static and dynamic graph learning models. The static baselines comprise the Brain Graph Benchmark using Principal Neighbourhood Aggregation (BrainGB-PNA)\cite{Cui2023BrainGB} and the Connectivity-based Graph Convolutional Network (cGCN)\cite{Wang2021GCNfMRI}. In contrast, the dynamic baselines include the Dynamic Brain Graph Deep Generative Model (DBGDGM)\cite{Campbell2023DBGDGM}, Dynamic Graph with Spatio-Temporal Transformer (DGST)\cite{Fan2023DGSTFormer}, Brain Tokenized Graph Transformer (TokenGT)\cite{Dong2023BrainTokenGT}, Structured Dynamic Graph Learning (SDGL)\cite{Li2023DynamicGraph}, Modularity-constrained Dynamic Representation Learning (MDRL)\cite{Wang2023Modularity}, Functional Brain Network Generator (FBNetGen)\cite{Kan2022FBNetGen}, and Joint Spatio-Temporal Graph Attention Network (JGAT)\cite{Chiu2023JGAT}.
\subsubsection{Experiment Settings}
We used 5-fold cross-validation and evaluated performance using accuracy (ACC), F1 score (F1), area under the curve (AUC), sensitivity (SEN), and specificity (SPE). Experiments were conducted on NVIDIA RTX 4070 using PyTorch.

\subsection{Performance Comparison}

In this section, we compare all graph learning methods on the ADNI, HCP, and ABIDE datasets.

We evaluate the proposed NeuroPathNet model on the ADNI dataset. We use a more challenging multi-class classification scenario, where the model needs to distinguish NC, MCI, and AD at the same time. Table \ref{table:multi-class} shows the comprehensive comparison results of different methods.  Our model shows consistent superiority in all evaluation indicators.
Specifically, the accuracy of NeuroPathNet reaches 66.67\%, which is 1.65\% higher than the second-ranked model cGCN (65.02\%), and the AUC score of 66.397\%  exceeds the second place JGAT (64.66\%), indicating stronger discrimination ability in multi-category scenarios. It is worth noting that our model achieves an F1 score of 75.14\%, indicating that our classification is balanced across all categories. The sensitivity (72.83\%) and specificity (73.20\%) metrics further verify the model's strong ability to simultaneously identify different cognitive states, which is actually quite difficult due to the complexity of brain maps.

The performance evaluation results of different models on the ABIDE dataset are presented in Table \ref{table:model-evaluation}. Our model demonstrates superior performance in autism spectrum disorder (ASD) classification across all evaluation metrics. Specifically, NeuroPathNet achieves an accuracy of 79.21\%, which is significantly higher than other baseline methods. This substantial improvement in accuracy indicates the effectiveness of our dynamic community-aware approach in capturing ASD-related brain network patterns.

The model also exhibits excellent performance in other key metrics. The AUC score of 78.12\% suggests strong discriminative ability between ASD and typically developing controls. The high F1 score (82.71\%) demonstrates balanced performance in terms of precision and recall. Particularly noteworthy are the sensitivity (83.10\%) and specificity (77.62\%) scores are crucial for clinical applications.

The evaluation results on the HCP cognition classification task are shown in Table \ref{table:hcp-emotion-classification}. The proposed model achieves state-of-the-art performance across all metrics, demonstrating its effectiveness in capturing cognition-related brain states. The model attains an AUC of 81.73\%, surpassing the second-best model cGCN (ACC: 80.63\%) by a margin of 1.10\%. 

Our model exhibits remarkable performance in discriminating different cognition states, as evidenced by the high AUC score of 83.58\%. This represents a substantial improvement over traditional approaches like DBGDGM (AUC: 73.34\%) and even recent advanced methods such as MDRL (AUC: 79.36\%). The balanced performance is further demonstrated by the F1 score of 81.91\%, which is comparable to cGCN (F1: 83.71\%) but significantly higher than other baseline methods. The sensitivity (80.03\%) and specificity (80.63\%) metrics indicate the model's robust ability to identify different cognitive states while maintaining low false positive rates.

\begin{table}[t!]
    \centering
    \caption{Performance comparison on ADNI dataset in multi-class classification task (\%)}
    \renewcommand{\arraystretch}{1.2}
    \setlength{\tabcolsep}{2pt}
    \begin{tabular}{|c|c|c|c|c|c|}
        \hline
        \textbf{} & \multicolumn{5}{c|}{\textbf{Evaluation Metrics}} \\
        \cline{2-6}
        \textbf{Method} & \textbf{ACC} & \textbf{AUC} & \textbf{F1} & \textbf{SEN} & \textbf{SPE} \\
        \hline
        DBGDGM & 52.77±6.81 & 54.59±4.96 & 65.93±7.05 & 56.29±5.70 & 52.73±8.19 \\
        \hline
        DGST & 57.38±7.95 & 56.21±6.88 & 68.94±6.31 & 55.08±7.65 & 58.49±5.76 \\
        \hline
        BrainGB & 55.70±6.59 & 58.79±5.19 & 67.82±7.23 & 57.01±7.03 & 74.94±8.12 \\
        \hline
TokenGT & 60.79±7.66 & 61.10±7.90 & 72.85±5.88 & 64.14±6.37 & 66.48±7.12 \\
        \hline
        SDGL & 61.06±6.29 & 58.46±6.59 & 70.40±7.35 & 58.74±5.62 & 61.11±8.30 \\
        \hline
        MDRL & 62.36±6.87 & 60.84±8.26 & 71.58±5.40 & 65.56±6.48 & 64.76±6.90 \\
        \hline
        FBNetGen & 64.51±8.78 & 62.04±6.14 & 73.53±7.36 & 64.73±6.07 & 63.40±6.17 \\
        \hline
        JGAT & 63.82±5.48 & 64.66±6.19 & 71.54±6.62 & 61.73±7.80 & 66.35±6.93 \\
        \hline
        cGCN & 65.02±6.28 & 63.26±9.73 & 73.77±4.95 & 66.46±7.61 & 65.26±6.17 \\
        \hline
        Ours & \textbf{66.67±9.77} & \textbf{66.39±5.74} & \textbf{75.14±5.18} & \textbf{72.83±6.19} & \textbf{73.20±7.68} \\
        \hline
    \end{tabular}
    \label{table:multi-class}
\end{table}

\begin{table}[b!]
    \centering
    \caption{Performance comparison on ABIDE dataset in classification task (\%)}
    \renewcommand{\arraystretch}{1.2}
    \setlength{\tabcolsep}{2pt}
    \begin{tabular}{|c|c|c|c|c|c|}
        \hline
        \textbf{} & \multicolumn{5}{c|}{\textbf{Evaluation Metrics}} \\
        \cline{2-6}
        \textbf{Method} & \textbf{ACC} & \textbf{AUC} & \textbf{F1} & \textbf{SEN} & \textbf{SPE} \\
        \hline
       DBGDGM & 71.66±7.82 & 73.74±6.43 & 70.93±8.91 & 74.82±5.67 & 72.51±7.34 \\
        \hline
       DGST & 72.43±6.78 & 74.31±8.23 & 71.24±5.46 & 75.53±7.91 & 73.32±6.34 \\
        \hline
        BrainGB & 74.25±8.34 & 76.57±5.78 & 73.06±7.23 & 77.11±6.89 & 74.94±8.12 \\
        \hline
        TokenGT & 73.55±5.92 & 75.85±7.45 & 72.33±6.78 & 76.83±8.34 & 74.19±5.67 \\
        \hline
       SDGL & 75.71±7.56 & 77.15±6.12 & 73.41±8.45 & 77.98±5.78 & 75.23±7.91 \\
        \hline
        MDRL & 72.65±1.70 & 65.60±2.10 & 57.00±2.80 & 74.10±3.10 & 65.60±1.90 \\
        \hline
        FBNetGen & 74.29±8.78 & 76.47±5.34 & 72.64±7.67 & 76.94±6.12 & 75.53±8.45 \\
        \hline
        JGAT& 75.56±5.45 & 77.01±7.89 & 73.33±6.34 & 77.25±8.67 & 76.18±5.23 \\
        \hline
        cGCN & 71.60±9.90 & 77.92±8.34 & 73.83±5.45 & 78.12±7.78 & 76.92±6.23 \\
        \hline
        Ours & \textbf{79.21±8.73} & \textbf{78.12±6.16} & \textbf{80.45±3.64} & \textbf{83.10±5.34} & \textbf{77.62±6.72} \\
        \hline
    \end{tabular}
    \label{table:model-evaluation}
\end{table}

\begin{table}[h!]
    \centering
    \caption{Performance comparison on HCP dataset in classification task (\%)}
    \renewcommand{\arraystretch}{1.2}
    \setlength{\tabcolsep}{2pt}
    \begin{tabular}{|c|c|c|c|c|c|}
        \hline
        \textbf{} & \multicolumn{5}{c|}{\textbf{Evaluation Metrics}} \\
        \cline{2-6}
        \textbf{Method} & \textbf{ACC} & \textbf{AUC} & \textbf{F1} & \textbf{SEN} & \textbf{SPE} \\
        \hline
        DBGDGM & 71.63±7.82 & 73.34±6.43 & 70.63±8.91 & 75.45±5.67 & 70.11±7.34 \\
        \hline
        DGST & 72.26±6.78 & 74.43±8.23 & 71.11±5.46 & 73.39±7.91 & 73.63±6.34 \\
        \hline
        BrainGB & 74.38±8.34 & 76.16±5.78 & 73.49±7.23 & 75.58±6.89 & 72.00±8.12 \\
        \hline
        TokenGT & 75.13±5.92 & 77.21±7.45 & 74.93±6.78 & 77.79±8.34 & 74.91±5.67 \\
        \hline
        SDGL & 76.43±7.56 & 78.69±6.12 & 75.75±8.45 & 79.13±5.78 & 76.20±7.91 \\
        \hline
        MDRL & 77.55±6.23 & 79.36±8.67 & 76.2±5.89 & 78.6±7.45 & 78.4±6.78 \\
        \hline
        FBNetGen & 73.65±8.78 & 80.10±5.34 & 74.43±7.67 & 76.81±6.12 & 71.76±8.45 \\
        \hline
        JGAT & 75.76±5.45 & 73.91±7.89 & 71.19±6.34 & 73.47±8.67 & 71.23±5.23 \\
        \hline
        cGCN & 80.63±6.89 & 78.21±8.34 & 81.31±5.45 & 79.03±7.78 & 76.90±6.23 \\
        \hline
        Ours & \textbf{81.73±9.84} & \textbf{86.26±7.69} & \textbf{83.71±8.30} & \textbf{80.03±5.94} & \textbf{80.63±6.91} \\
        \hline
    \end{tabular}
    \label{table:hcp-emotion-classification}
\end{table}

\subsection{Parameter Analysis}
In order to optimize the model performance, we conducted a grid search for key hyperparameters. We used 32 channels in the network structure to enhance the model's modeling capabilities in spatial and temporal dimensions. At the same time, in order to effectively control overfitting while improving the model's expressiveness, we set a low Dropout ratio (0.1) and a moderate Batch Size (32) to maintain the generalization ability of the model.

In addition, we also focused on tuning the structural depth of the graph convolutional network. Finally, a 6-layer graph convolution structure was adopted to avoid problems such as gradient vanishing or training instability caused by the network being too deep. The learning rate was set to 0.1, which can speed up the convergence speed in the initial training stage and also showed good stability in the experiment.

\subsection{Ablation Study}

\subsubsection{The difference between different classification methods}

\begin{table}[b!]
\centering
\caption{Performance comparison using different static functional partitioning strategies (\%)}
\renewcommand{\arraystretch}{1.2}
\setlength{\tabcolsep}{5pt}
\begin{tabular}{|c|c|c|c|c|c|c|}
\hline
Method & \#Partitions & ACC & AUC & F1 & SEN & SPE \\
\hline
Yeo-7  & 7 & 76.30 & 78.60 & 76.50 & 73.05  & 77.94\\
Yeo-17 & 17 & 78.88 & 83.12 & 80.76 & 77.92 & 76.33 \\
Schaefer-100 & 10  &81.73 & 86.26 & 83.71 & 80.03 & 80.63 \\
\hline
\end{tabular}
\label{table:partition_methods_evaluation}
\end{table}

In order to study the impact of different static functional partitioning strategies on model performance, we selected three brain region partitioning methods that are widely used in brain network research: Yeo-7, Yeo-17, and Schaefer-100. The Yeo partitioning scheme uses clustering to partition resting-state fMRI data, where Yeo-7 divides the cerebral cortex into 7 functional communities, and Yeo-17 further refines it into 17 functional communities. In contrast, the Schaefer partitioning scheme provides a higher spatial resolution partitioning method, dividing the cerebral cortex into 100 regions (Schaefer-100), and clustering these regions into 10 functional networks based on gradients and global similarities. 

In order to evaluate the impact of different static functional partitioning strategies on model performance, we conducted comparative experiments on three commonly used partitioning methods under the NeuroPathNet framework. The results are shown in Table~\ref{table:partition_methods_evaluation}. The Yeo-7 partitioning scheme provides seven coarse-grained brain region functional modules and achieved relatively good results; Yeo-17 improved the F1 value and accuracy through finer-grained partitioning. In contrast, the method using Schaefer-100 partitioning (aggregated into 10 functional networks) performed best, indicating that this partitioning scheme provides better spatial granularity while maintaining functional consistency, which helps the model learn the interactive features of different trajectories more effectively.

\subsubsection{Analysis on Different Modules}
\begin{table}[b!]
\centering
\caption{Ablation study on four core modules of NeuroPathNet (\%)}
\renewcommand{\arraystretch}{1.2}
\setlength{\tabcolsep}{5pt}
\begin{tabular}{|c|c|c|c|c|c|c|c|c|}
\hline
PM & GE & MHA & TP & ACC & AUC & F1 & SEN & SPE\\
\hline
\checkmark &  &  &  & 71.41 & 72.71 & 70.12 & 66.81 & 68.41 \\
\checkmark & \checkmark &  &  & 76.28 & 78.51 & 75.80 & 71.38 & 73.94 \\
\checkmark & \checkmark & \checkmark &  & 80.11 & 81.79 & 79.26 & 75.71 & 78.03 \\
\checkmark & \checkmark & \checkmark & \checkmark & \textbf{81.73} & \textbf{86.26} & \textbf{83.71} & \textbf{80.03} & \textbf{80.63} \\
\hline
\end{tabular}
\label{table:core_module_ablation}
\end{table}

In order to evaluate the role of each core module in the NeuroPathNet framework, we conducted an ablation experiment of gradually adding modules, introducing path modeling (PM), global integration (GE), multi-head attention mechanism (MHA) and time pooling (TP) in turn. As shown in Table~\ref{table:core_module_ablation}, it can be seen that with the gradual addition of modules, the model performance continues to improve in various indicators. The use of path modeling can provide basic dynamic trajectory connection modeling capabilities. The addition of global integration can enhance the mutual influence between different paths. The multi-head attention mechanism strengthens the learning ability. After the introduction of time pooling, the model achieves optimal performance. The experiment verifies the complementary advantages of local modeling and global dynamic integration, and proves the effectiveness of the model.

\subsection{Visualization} 

\begin{figure}[t!]
    \centering
    \begin{subfigure}[b]{0.15\textwidth}
        \centering
        \includegraphics[width=\textwidth]{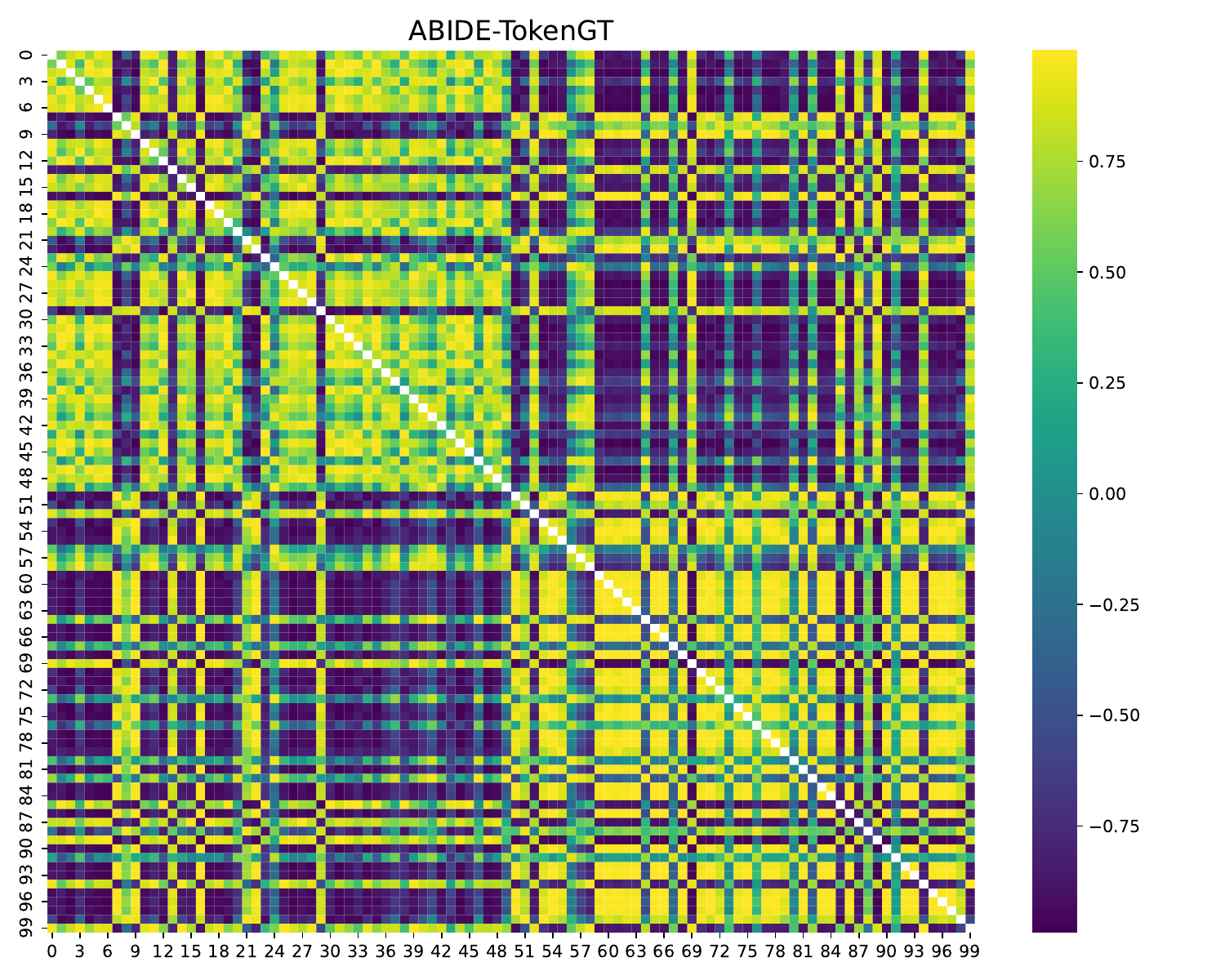}
        \caption{ABIDE-TokenGT}
        \label{fig:tsne1}
    \end{subfigure}
    \begin{subfigure}[b]{0.15\textwidth}
        \centering
        \includegraphics[width=\textwidth]{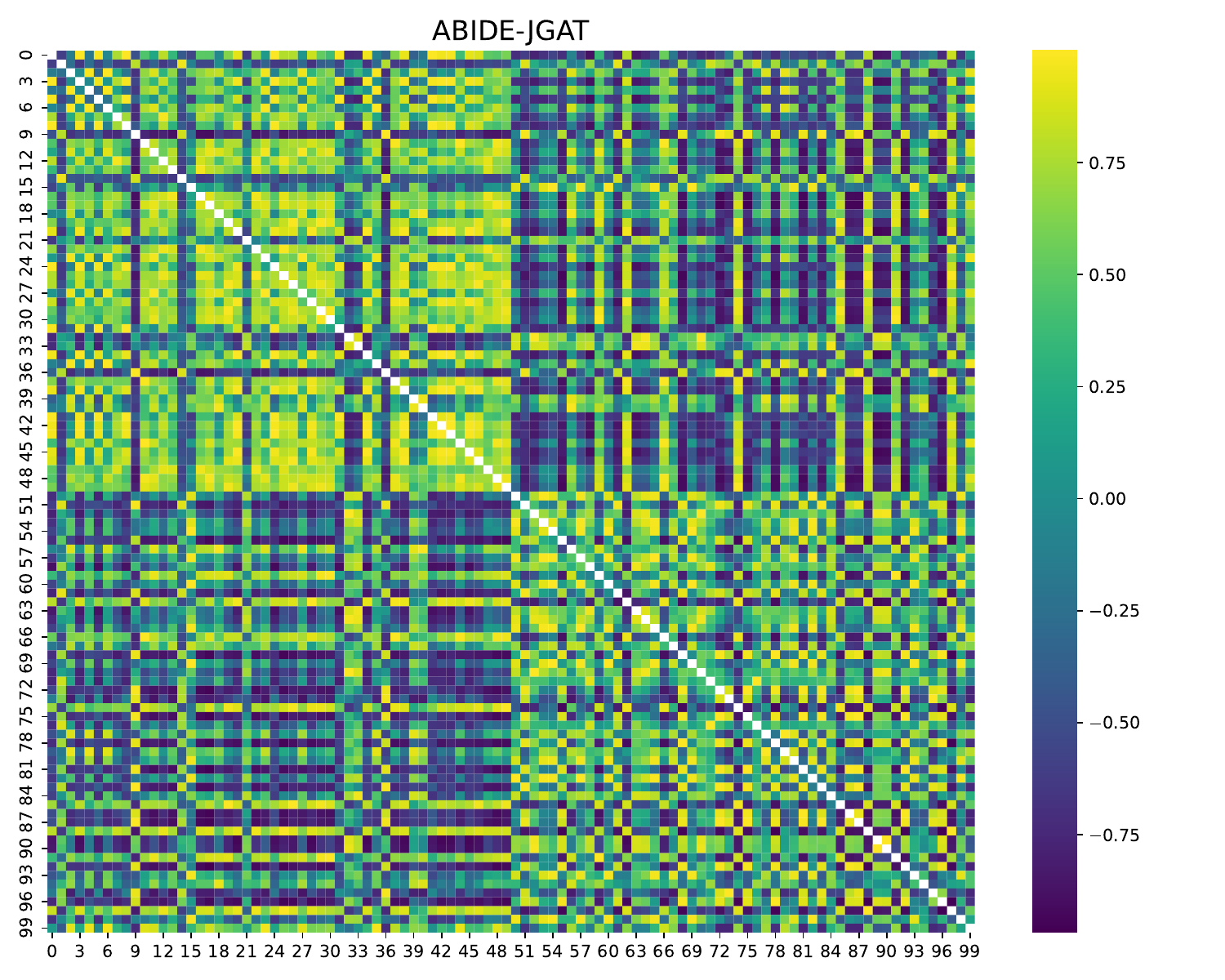}
        \caption{ABIDE-JGAT}
        \label{fig:tsne2}
    \end{subfigure}
    \begin{subfigure}[b]{0.15\textwidth}
        \centering
        \includegraphics[width=\textwidth]{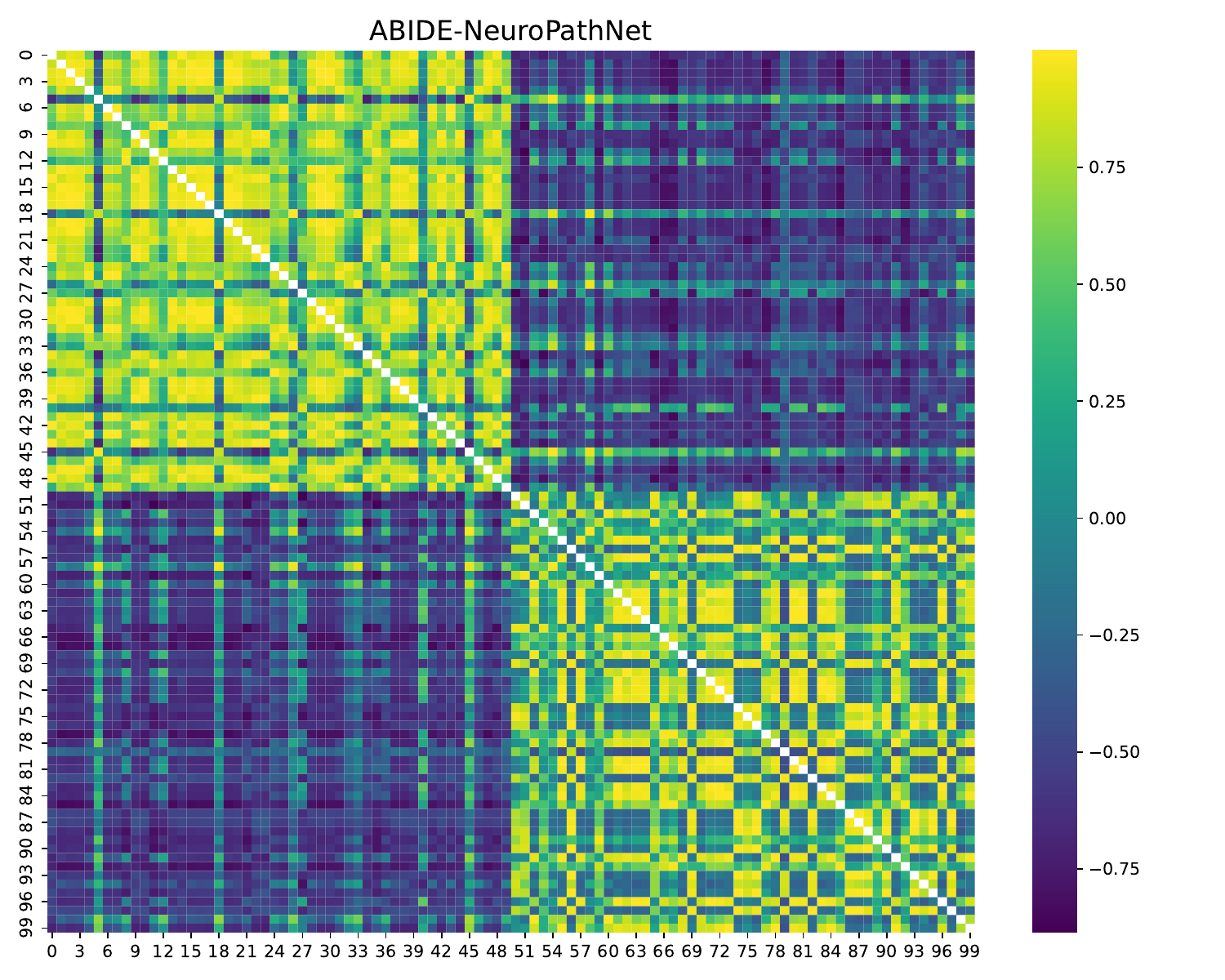}
        \caption{ABIDE-Ours}
        \label{fig:tsne3}
    \end{subfigure}
    
    \begin{subfigure}[b]{0.15\textwidth}
        \centering
        \includegraphics[width=\textwidth]{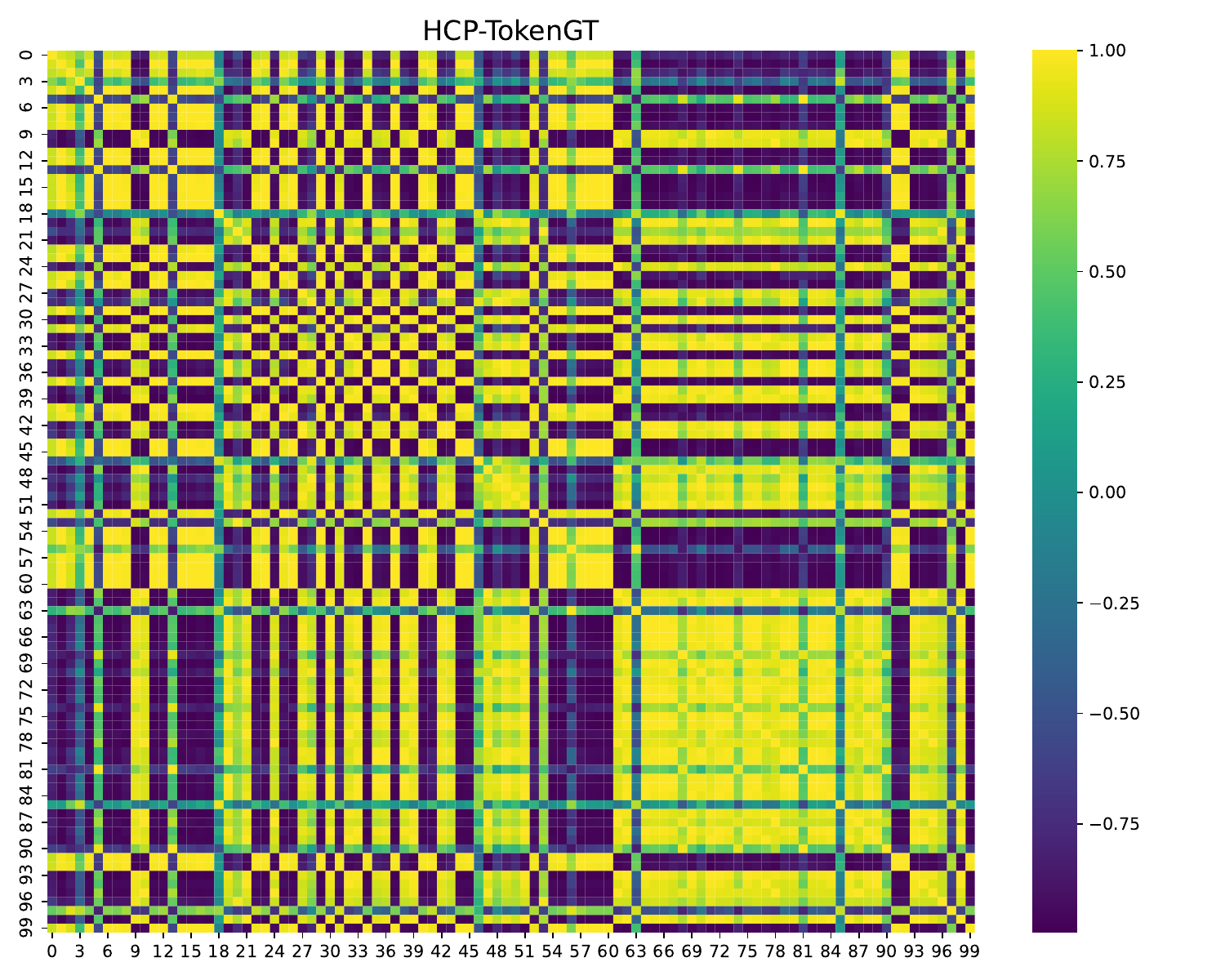}
        \caption{HCP-TokenGT}
        \label{fig:tsne4}
    \end{subfigure}
    \begin{subfigure}[b]{0.15\textwidth}
        \centering
        \includegraphics[width=\textwidth]{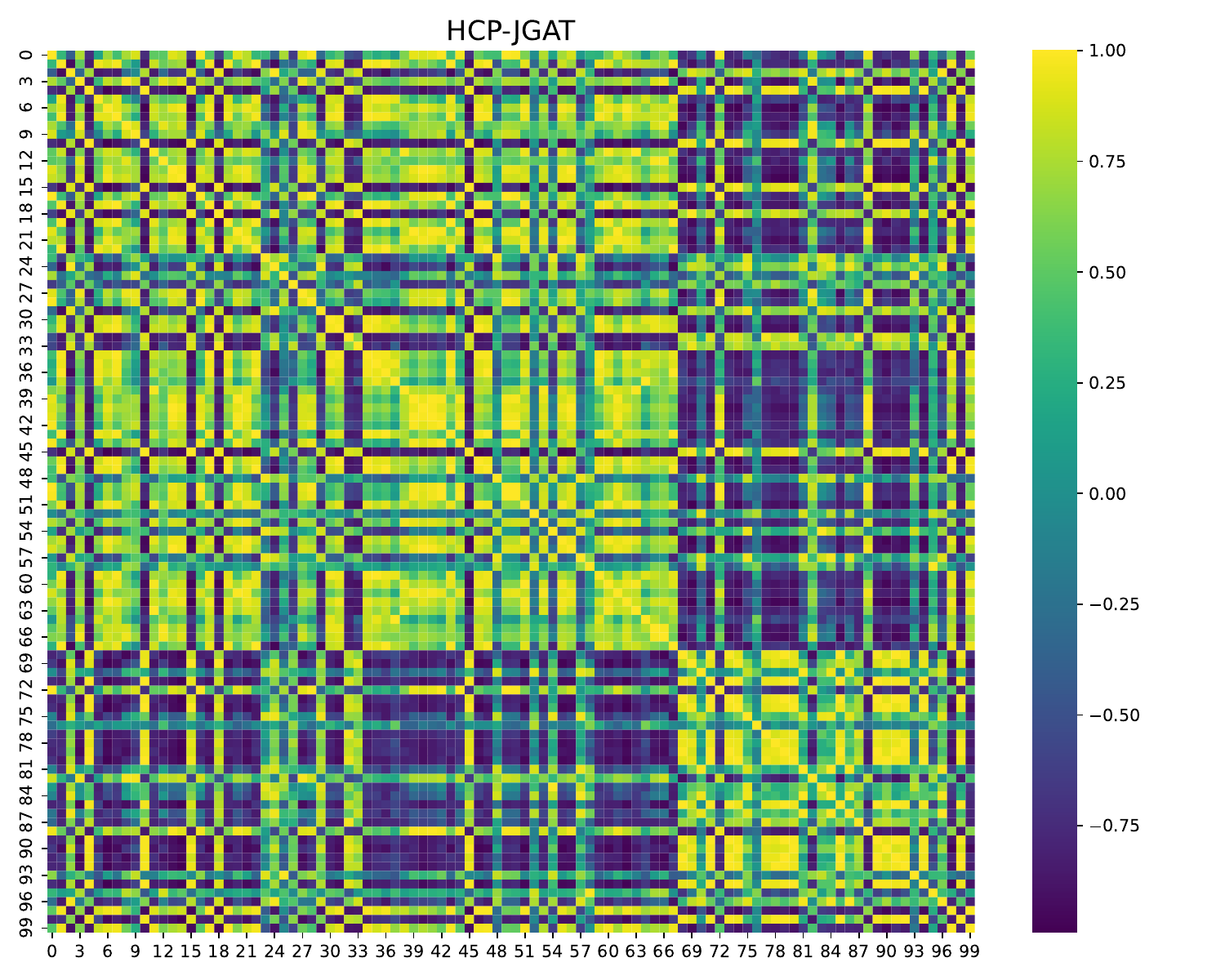}
        \caption{HCP-JGAT}
        \label{fig:tsne5}
    \end{subfigure}
    \begin{subfigure}[b]{0.15\textwidth}
        \centering
        \includegraphics[width=\textwidth]{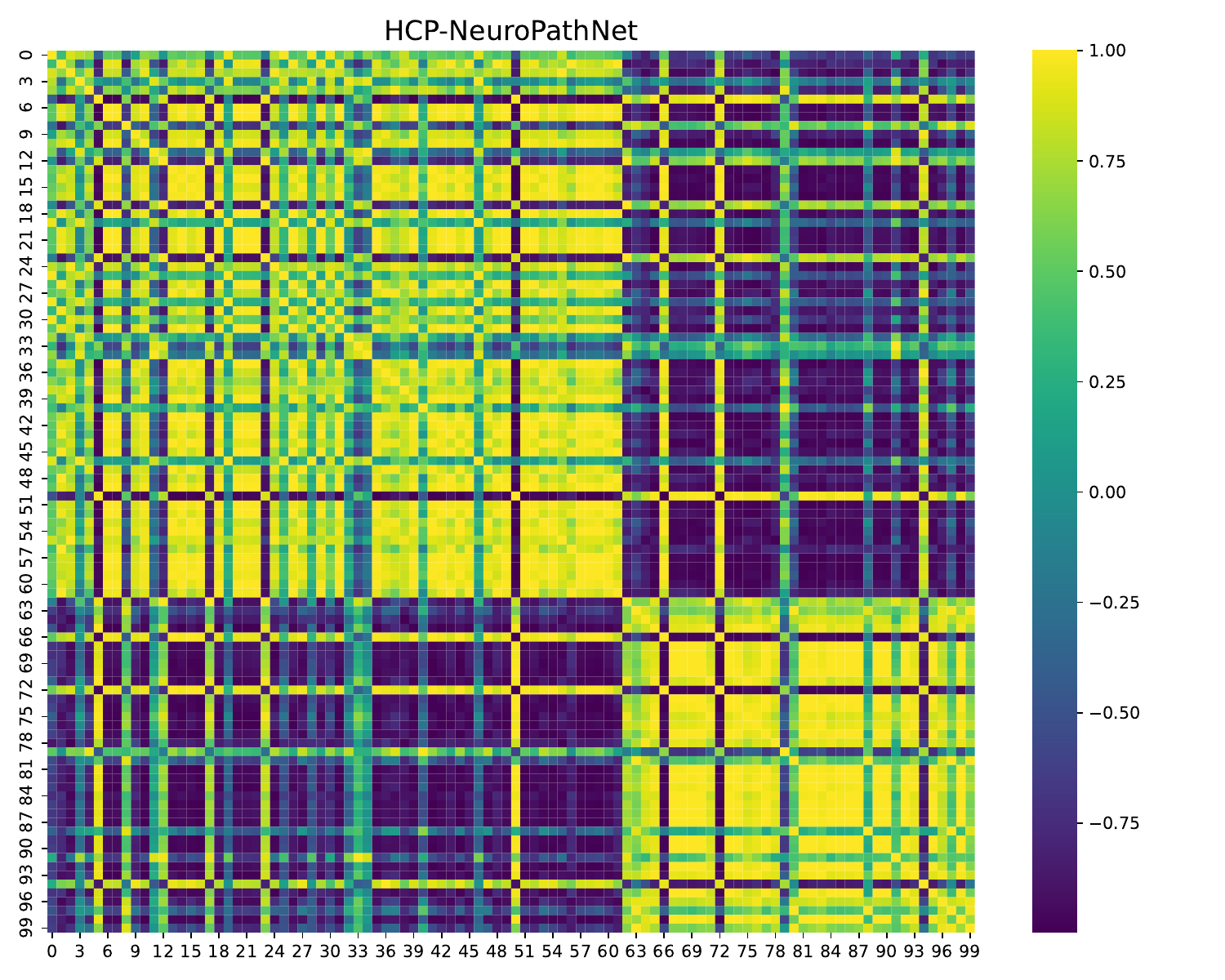}
        \caption{HCP-Ours}
        \label{fig:tsne6}
    \end{subfigure}

    \caption{The similarity matrices of sample representations learned by TokenGT, Graphormer, and NeuroPathNet}
    \label{fig:simi}
\end{figure}
The similarity matrices of sample representations learned by TokenGT, Graphormer, and NeuroPathNet were calculated and visualized respectively. The results are shown in Fig. \ref{fig:simi}. The closer the color is to yellow, the more similar their classification results are; the closer it is to blue, the more likely they belong to different categories. From the similarity matrix of TokenGT (Fig. \ref{fig:tsne1}, Fig. \ref{fig:tsne4}), it can be seen that the boundaries between different blocks are relatively fuzzy, and the classification results are not prominent enough. The similarity matrix of the JGAT model (Fig. \ref{fig:tsne2}, Fig. \ref{fig:tsne5}) still has many transition areas and is uncertain when processing samples. The NeuroPathNet model proposed in this paper (Fig. \ref{fig:tsne3}, Fig. \ref{fig:tsne6}) has the strongest learning effect. The inter-class regions of different categories appear darker blue, and the model can distinguish different categories well. The visualization results prove that NeuroPathNet obtains more effective information.

\section{Conclusion}

In this paper, we divided the human brain into multiple static functional partitions, and there are complex and dynamic interactions between these partitions. To further study these interactions, we propose a novel spatiotemporal-aware neural network analysis framework, NeuroPathNet. This model innovatively introduces the idea of path-level modeling, takes functional partitions as the basic mapping unit, and describes the collaborative patterns between different functional modules that evolve over time by constructing and analyzing functional connection trajectories across partitions. Experimental results on three public datasets showed that NeuroPathNet significantly outperformed existing static and dynamic methods in multiple neurological disease classification tasks. Future work will explore multimodal neuroimaging fusion (EEG, MEG, DTI) and extend this method to broader neurological disease detection tasks.



\ifCLASSOPTIONcaptionsoff
  \newpage
\fi

\bibliographystyle{IEEEtran}

\bibliography{refs.bib}

\end{document}